\documentclass{article}

\usepackage{arxiv}

\usepackage[utf8]{inputenc} 
\usepackage[T1]{fontenc}    
\usepackage{url}            
\usepackage{booktabs}       
\usepackage{nicefrac}       
\usepackage{microtype}      
\usepackage{xcolor}         

\usepackage{amsmath,amsfonts,bm}

\def\eqref#1{equation~\ref{#1}}

\def\1{\bm{1}}

\DeclareMathAlphabet{\mathsfit}{\encodingdefault}{\sfdefault}{m}{sl}
\SetMathAlphabet{\mathsfit}{bold}{\encodingdefault}{\sfdefault}{bx}{n}

\def\gH{{\mathcal{H}}}

\def\gS{{\mathcal{S}}}
\def\gT{{\mathcal{T}}}

\newcommand{\E}{\mathbb{E}}

\DeclareMathOperator*{\argmin}{arg\,min}

\newcommand{\rbr}[1]{\left(#1\right)}
\newcommand{\sbr}[1]{\left[#1\right]}
\newcommand{\cbr}[1]{\left\{#1\right\}}

\def\bb{\begin{equation}} \def\ee{\end{equation}}

\usepackage{hyperref}       
\usepackage{svg}
\usepackage{multirow}
\usepackage{float}
\renewcommand{\eqref}{Eq.~\ref}
\newcommand{\tabref}{\textbf{Table}.~\ref}
\usepackage[linesnumbered,ruled,vlined]{algorithm2e}
\usepackage[numbers]{natbib}

\newcommand{\grad}{\nabla}

\newcommand{\emb}{\texttt{emb}}

\title{Graph Condensation via Receptive Field Distribution Matching}

\author{%
  Mengyang Liu \\
  Georgia Institute of Technology\\
   \And
   Shanchuan Li \\
   Georgia Institute of Technology \\
   \AND
   Xinshi Chen \\
   Georgia Institute of Technology \\
   \And
   Le Song \\
   Georgia Institute of Technology \\
}

\begin{document}

\maketitle

\begin{abstract}
  Graph neural networks (GNNs) enable the analysis of graphs using deep learning, with promising results in capturing structured information in graphs. This paper focuses on creating a small graph to represent the original graph, so that GNNs trained on the size-reduced graph can make accurate predictions. We view the original graph as a distribution of receptive fields and aim to synthesize a small graph whose receptive fields share a similar distribution. Thus, we propose \textbf{G}raph \textbf{C}ondesation via \textbf{R}ceptive \textbf{F}ield \textbf{D}istribution \textbf{M}atching (GCDM), which is accomplished by optimizing the synthetic graph through the use of a distribution matching loss quantified by maximum mean discrepancy (MMD). Additionally, we demonstrate that the synthetic graph generated by GCDM is highly generalizable to a variety of models in evaluation phase and that the condensing speed is significantly improved using this framework.
\end{abstract}

\section{Introduction}

Many real-world datasets, including social networks, molecular interactions, and recommendation systems are graph-structured. Recently, a substantial line of research on graph neural networks (GNNs) enables the analysis of graphs using deep learning, with promising results in capturing structured information in graphs. 

Numerous real-world graphs are large-scale, containing millions of nodes and trillions of edges. Due to the non-euclidean nature of the network and the complicated dependencies between nodes, training a GNN over a large graph can be very expensive. More specifically, we often need to train a GNN multiple times on the same graph to validate the design choices made when tuning the hyperparameters, searching the architecture, and so on, or when considering the continual learning scenario. As a consequence, there is a considerable interest in techniques that reduce the computational cost of training multiple GNNs on the same graph without losing the performance.

To address these issues, this paper focuses on creating a small but informative graph to represent the original graph, so that GNNs trained on the size-reduced graph can make accurate predictions on the original graph.  

Graph sparsification and graph coarsening are two well-known techniques for graph simplification. By reducing the number of edges, graph sparsification attempts to mimic a graph with a more sparse graph, whereas graph coarsening directly reduces the number of nodes (using a subset of the original node set). While the motive for the two techniques is self-evident, the disadvantage is equally self-evident: (1) sparsification has a diminishing effect when the graph's complexity is primarily derived from node features. (2) Both methods seek to maintain a spectral attribute of the original graph, such as the principal eigenvalues of the Laplacian Matrix, although this may not be the ideal option due to the possibility of preserving significant noise.

\cite{jin2021graph} recently proposed a deep learning based method called GCOND. Its central concept is \textit{synthesizing} a small graph, by minimizing the \textit{gradient matching} loss between the gradients of training losses w.r.t. GNN parameters given by the original graph and the synthetic graph. While GCond has demonstrated advantages over traditional approaches, it has two downsides. Firstly, while minimizing the gradient matching loss, the condensation procedure is expensive due to the need of computing the second-order derivative w.r.t. GNN parameters. For example, it takes about 100 minutes (running on a single RTX8000 GPU) to condense the Ogbn-arxiv dataset to 1\% by 50 epochs. Secondly, since the gradient matching loss is architecture-dependent, the condensed graph may not generalize well to new GNN architectures.

\begin{figure*}[htbp]
    \centering
    \includegraphics[width=1\textwidth]{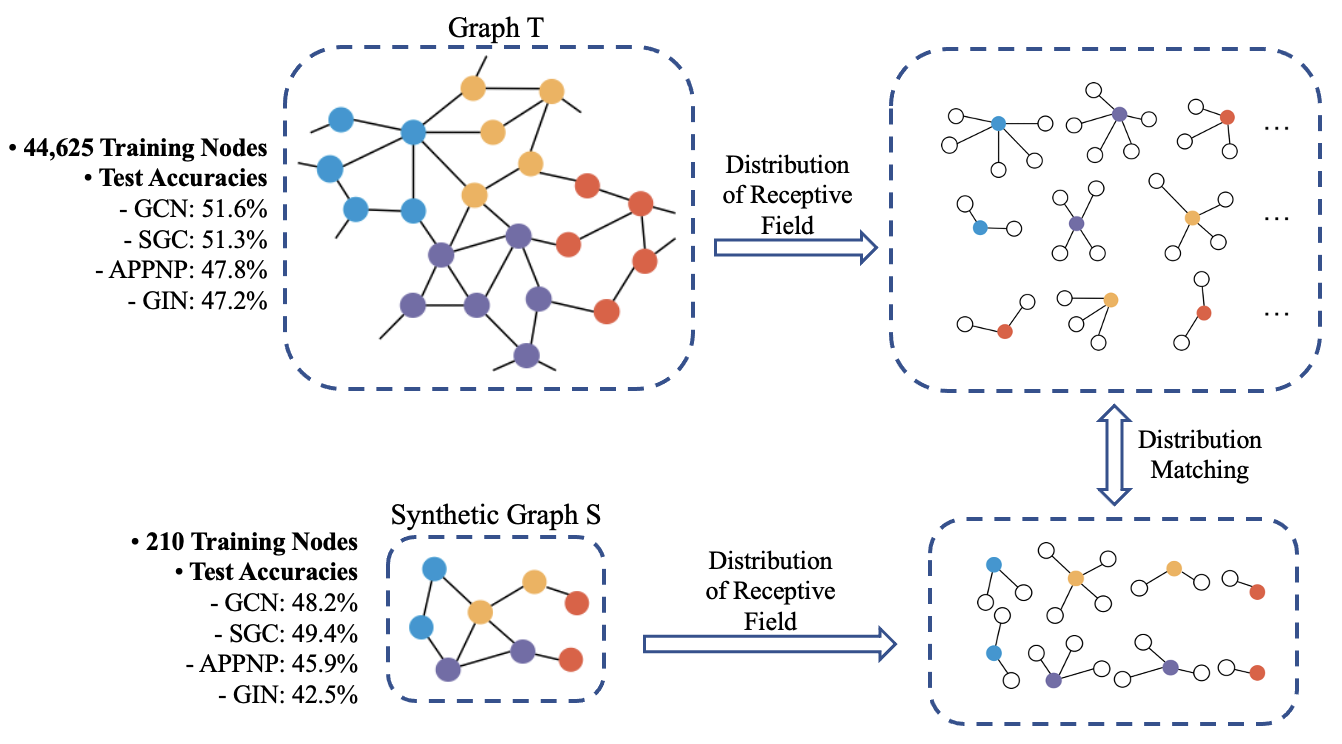}
    \caption{The overall framework of graph condensation via receptive field distribution matching (GCDM) and the test performance on Flickr dataset with 99.5\% size reduction.}
    \label{fig:framework}
\end{figure*}

To address the aforementioned difficulty, we present Graph Condensation through receptive field Distribution Matching (\textbf{GCDM}). More precisely, we view the original graph as \textit{a distribution of receptive fields} and aim to synthesize a small graph whose receptive fields share a similar distribution as the original graph (see Figure~\ref{fig:framework}). This is accomplished by optimizing the synthetic graph through the use of a distribution matching loss quantified by maximum mean discrepancy (MMD). In contrast to GCOND, our condensation loss does not rely on the training loss of optimizing parameters of a specific GNN. Therefore, the condensed graph can be used for training various GNN models. Furthermore, it avoids computing the second-order gradient w.r.t. the GNN parameters, which reduces the cost of the condensation process.

The use of distribution matching method for dataset condensation has been proposed in a recent Arxiv paper by~\cite{zhao2021dataset}, which is applied to image dataset condensation. However, applying the distribution matching method to condense graph-structured datasets is more challenging due to the complex dependencies between the nodes. As an example, it is straightforward to define a data distribution given an image dataset since each image can be viewed as an independent sample. However, in a graph-structured dataset, the nodes and edges are no longer independent. We need to carefully utilize the concept of receptive field to define the distribution that we aim to match. We will introduce the details in Sec~\ref{sec:method}.

To summarize, the main contributions of our work are:
\begin{itemize}
  \item \textit{Methodology:} We present a novel method for graph condensation, in which we define a distribution matching problem based on the concept of receptive fields.
  \item \textit{Accuracy:} Experimentally, we demonstrate the effectiveness of  GCDM on a variety of graph datasets. GNNs trained on small graphs produced by GCDM can achieve comparable performances to those trained on the original graphs. For instance, GCDM can approximate the original test accuracy by 97.6\% on Filckr, 90.6\% on Ogbn-arxiv with a 99\% graph size reduction, and it can approximate 99\% test accuracy on Cora, CiteSeer and PubMed with a 99\% graph size reduction.
    \item \textit{Generalization ability:} Our approach is broadly applicable to all GNN models, including generative invertible networks (GINs) \cite{gin2018}, which excel in discovering pathophysiologic information. For example, we achieved accuracy of 42.2\% (GCDM-X) and 42.5\% (GCDM) on the Flickr dataset with GIN, compared to the benchmark approach GCOND's \cite{jin2021graph} 29.5\% (GCOND-X) and 38.8\% (GCOND).
    \item \textit{Efficiency:} Our method is more efficient than the present graph condensation counterpart. For instance, in order to generate a 1\% synthetic graph for 50 epochs, GCDM takes an RTX8000 instance 1,782 GPU seconds while GCOND takes the same instance 5,960 GPU seconds.
\end{itemize}

\newpage
\section{Related Work}
\subsection{Graph Neural Networks.} 
Graph neural networks (GNNs) have gained increasing popularity in both research and applications due its powerful representability on graph-structured data, which enables the use of neighbor information to generate more expressive representations. \cite{compgnn2021, kipf2017semisupervised, gin2018, klicpera2019predict, hamilton2018inductive, tang2019chebnet}. The methodology has been implemented in various real-world applications such as recommender systems \cite{wang2021graph}, computer vision \cite{nazir2021survey}, and natural language processing \cite{wu2021graph}. 

\subsection{Coreset \& Data Condensation} Coresets are small and highly informative weighted subsets of the original dataset that can be used to approximate its performance. Numerous works have studied the selection strategy to facilitate the efficient training of deep learning models in supervised learning scenarios. Methods include maximizing the diversity of selected sample using parameters gradients as the feature \cite{aljundi2019gradient}, approximation of the coreset and the original data distributions \cite{DBLP:journals/corr/abs-1203-3472, rebuffi2017icarl, belouadah2020scail, campbell2019automated}, incrementally choosing data point with the largest negative implicit gradient \cite{borsos2020coresets}. However, the coreset selection is upper-bounded by the existing features, which could potentially constrain its expressiveness. Data condensation was proposed to address the aforementioned issue since it seeks to generate new samples that incorporate more information. Although a recently proposed gradient matching solution to data condensation seems to achieve promising performance \cite{DBLP:conf/icml/ZhaoB21}, it may provide difficulties since it was designed for image data.

\subsection{Graph Coarsening \& Graph Sparsification.}  Similar to coreset, graph coarsening is the idea to reduce the size of graph data while keeping its basic properties.\cite{loukas2018graph} considered the graph reduction problem from the perspective of restricted spectral approximation while modifying the measure utilized for the construction of graph sparsifiers. Graph sparsification \cite{spielman2010spectral}, on the other hand, tries to make a graph sparser while only removing edges. While coreset, graph coarsening, and graph sparsification methods are generally computationally efficient, they do come with their limitations. All three methods conduct greedy and incremental searches. Such searches don't guarantee global optimal. Additionally, coreset and graph coarsening both depend on the assumption that there exist highly representative samples in the original data; graph sparsification, on the other hand, makes the assumption that a large number of insignificant edges that can be removed. Both of the preceding are not always true.

\subsection{Data Condensation \& Graph Condensation.}
In contract to data selection methods such as coreset, the goal of data condensation \cite{wang2020dataset, DBLP:conf/icml/ZhaoB21} is to synthesize informative samples rather than selecting from existing samples. Recently, \cite{zhao2021datasetgm} proposed matching the gradients w.r.t a neural network weights between the original data and the synthetic data. Then, \cite{zhao2021datasetdm} further reduced the computational cost by matching the distribution of the two datasets. Similar to data condensation, graph condensation seeks to generate a synthetic graph that resembles the original data. \cite{jin2021graph} recently attempted to match the network gradients of GNNs and generate an adjacency matrix for the synthetic graph using a trained multilayer perceptron. It achieved similar performance to the original graph. However, this method has limitations in terms of generalizability and computational speed.

\section{Methodology}
\label{sec:method}

This section introduces our \textit{ graph condensation via receptive field distribution matching} framework, GCDM. We begin with a brief review of the graph condensation problem and existing approaches in Sec~\ref{sec:graph-condensation-problem}, and then discuss our methodology in greater detail in the followed sub-sections.
\subsection{Graph Condensation Problem}
\label{sec:graph-condensation-problem}
Consider a graph dataset $\mathcal{T} = \{A, X, Y\}$ where $A \in \mathbb{R}^{N\times N}$ is an adjacency matrix, $X \in \mathbb{R}^{N \times d}$ is a matrix of node features, and  $Y \in \{0, ..., C-1\}^N$ denotes the label of nodes. The constants $N,\ d,\ C$ are the number of nodes, the dimension of node features, and the number of classes respectively. Graph condensation aims to find a small graph $\mathcal{S} = \{A', X', Y'\}$ with $A' \in \mathbb{R}^{N' \times N'}, X' \in \mathbb{R}^{N' \times d}, Y' \in \{0,...,C-1\}^{N'}$ and $N' \ll N$ such that GNNs trained on $\mathcal{S}$ can achieve similar performances to GNNs trained on the original graph $\mathcal{T}$. Therefore, this problem can be formulated as the following bi-level optimization problem,

\begin{equation}\label{eq:Obj}
\begin{split}
    &\min_{\mathcal{S}}\mathcal{L}(\Psi_{\theta^{\mathcal{S}}}(A, X), Y) \\
    s.t. \ \  &\theta^{\mathcal{S}} = \argmin_{\theta} \mathcal{L}(\Psi_{\theta}(A', X'), Y') 
\end{split}
\end{equation}

where $\Psi_{\theta}$ denotes a GNN parameterized by $\theta$, $\mathcal{L}$ denotes the training loss function (i.e cross-entropy), and $\theta^{\mathcal{S}}$ denotes the set of parameters obtained by minimizing the training loss on the small graph $\mathcal{S}$.


\textbf{Existing Solutions.} Previous work on data condensation attempted to solve \eqref{eq:Obj} by matching the gradient of the neural network~\cite{zhao2021datasetgm}, and \cite{jin2021graph} further extended this method to graph neural networks. The gradient matching objective in \cite{jin2021graph} is

\begin{equation}
    \label{eq:gmgraph}
    \min_{\mathcal{S}} E_{\theta_0 \sim P_{\theta_0}} [\sum_{t=0}^{T-1} D(\nabla_{\theta} \mathcal{L}(\Psi_{\theta_t}(A', X'), Y'), \nabla_{\theta} \mathcal{L}(\Psi_{\theta_t}(A, X), Y))]
\end{equation}
where $P_{\theta_0}$ is a distribution of parameter initialization of $\Psi$, and $D$ is a distance measurement.

\textbf{Challenges.} Optimizing \eqref{eq:gmgraph} can result in a condensed graph that produces similar gradients as the original graph during training. However, the gradients are only guaranteed to be similar on the particular model structure $\Psi_{\theta}$ used during the condensation. The generalization ability of training other GNN architectures on the condensed graph is not guaranteed by the objective in \eqref{eq:gmgraph}.
Additionally, optimizing \eqref{eq:gmgraph} could be very expensive because it involves computing the second order derivative w.r.t. the parameters in $\Psi$. In the subsequent experiment part, we will also compare the efficiency of our method to that of the gradient matching method.

\textbf{Our Solution: Distribution Matching between Graphs.} To overcome the above-mentioned challenges, we propose Graph Condensation via Distribution Matching ({GCDM}). This method is inspired by \cite{zhao2021dataset} which proposed to condense image datasets via distribution matching. The main idea in \cite{zhao2021dataset} is to synthesize a small set of images  $\mathcal{S}_{IMG}$  that can accurately approximate the data distribution of real images $\mathcal{T}_{IMG}$, by minimizing the maximum mean discrepancy (MMD) between them:
\begin{align}
    \min_{\mathcal{S}_{IMG}} \text{MMD}(\mathcal{S}_{IMG}, \mathcal{T}_{IMG}).
\end{align}
The method of distribution matching can be more advantages because it \textit{only compares the data distributions induced by the two datasets}, without relying on the configurations in the training process. However, migrating the idea of distribution matching to condense graph-structured dataset $\mathcal{T}$ is not straightfoward. It is unclear how to view a graph $\mathcal{T}$ as a data distribution since the nodes in the graph are connected by edges, unlike images which are viewed as independent samples. In the following, we will introduce how we formulate the graph condensation problem as a distribution matching problem.

\subsection{Viewing a Graph as a Distribution}


Before going into the details of our proposed GCDM, we first introduce \textit{how a graph $\mathcal{T}$ is converted to a distribution} under the context of node classification tasks. 

\subsubsection{Node classification by GNN}
For a node classification dataset $\mathcal{T}=\{A,X,Y\}$, training a GNN $\Psi_\theta$ is accomplished by optimizing the classification loss over the nodes
\begin{align}
   \mathcal{L}(\Psi_\theta(A,X), Y) & = \frac{1}{N}\sum_{i=1}^N \ell \rbr{\psi_\theta(A, X, i ), Y_i} 
    \label{eq:loss}
    \\ & = \E_{\rbr{(A, X, i ), Y_i }\sim P_\gT} \sbr{ \ell \rbr{\psi_\theta(A, X, i ), Y_i}  },\label{eq:loss2}
\end{align}
where $\ell$ is a loss function, $\psi_\theta(A, X, i )$ is the class predicted by the GNN $\Psi_\theta$ for the $i$-th node, and $Y_i$ is the $i$-th label. According to \eqref{eq:loss} and \eqref{eq:loss2}, we can view each pair $\rbr{(A, X, i ), Y_i }$ as an independent data point, and view the training graph as an empirical distribution $P_\gT$ induced by these data points:
\begin{align}
   \cbr{\rbr{(A, X, 1 ), Y_1 }, \rbr{(A, X, 2), Y_2 },\cdots, \rbr{(A, X, N ), Y_N } }.
\end{align}
However, representing and matching such a distribution seem to be complicated, because each data point involves a large adjacency matrix $A$. Therefore, we make use of the concept of receptive field to simplify this distribution.

\subsubsection{Receptive Fields of GNN} 
\begin{figure}[h!]
    \centering
    \includegraphics[width=\linewidth]{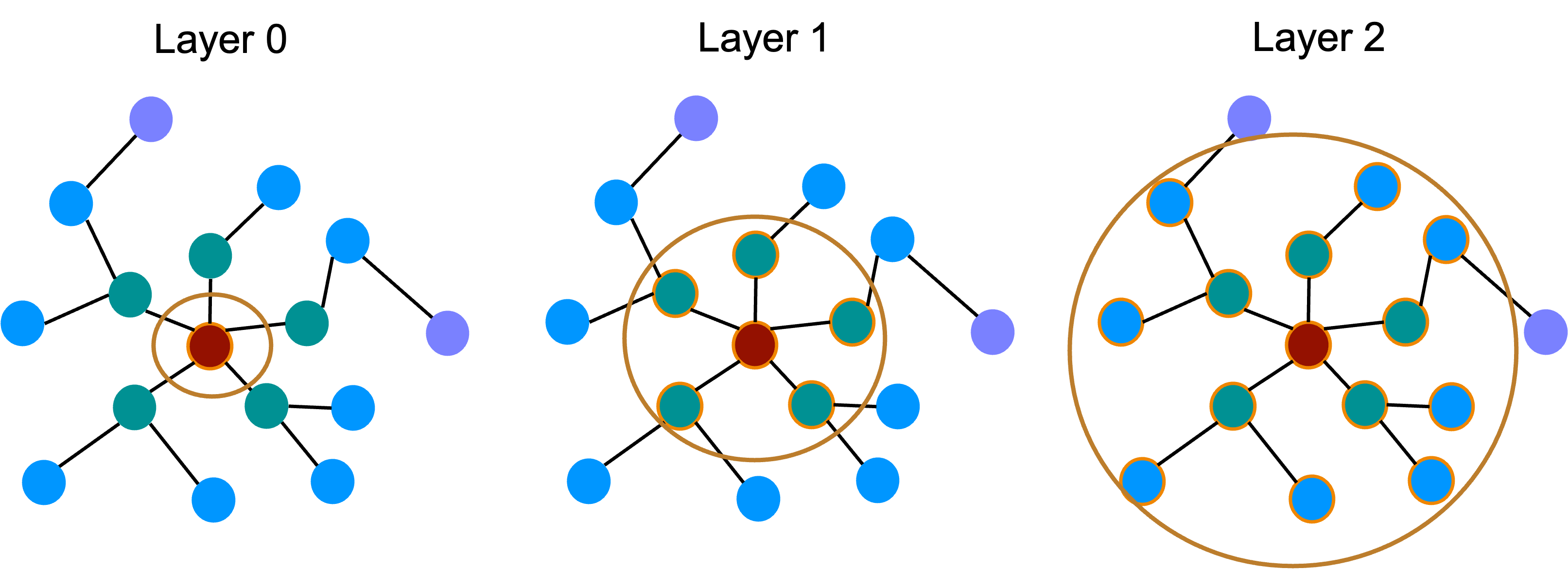}
    \caption{Receptive field $R(i, L)$ of GNNs for a node $i$ (red-colored) with $L=0, 1,2$.}
    \label{fig:receptive_field}
\end{figure}

In \eqref{eq:loss}, the label of node $i$ is predicted by $\psi_\theta(A, X, i )$. In fact, most GNNs only makes use of a \textit{local graph} around node $i$ to compute $\psi_\theta(A, X, i )$. As an example, for a $L$-layer Graph Convolutional Neural Network (GCN), only the information of the target node's $L$-hop neighbors contributes to the representation and prediction, so as other message-passing based GNNs regardless of the specific architecture. In the literature of GNN researches, a target node's $L$-hop local graph is called the \textit{receptive field} of a $L$-layer GNN (See Figure~\ref{fig:receptive_field}).

In this paper, we use the notation $R(i,L)$ to represent the receptive field of a $L$-layer GNN for node $i$. Clearly, if $L$ is larger the the diameter of graph, then $R(i,L)$ contains the whole graph.



\subsubsection{Viewing a Graph as a Distribution of Receptive Fields}  

Given the concept of receptive fields $R(i,L)$, we can rewrite the loss in \eqref{eq:loss} as
\begin{align}
   \mathcal{L}(\Psi_\theta(A,X), Y) & = \frac{1}{N}\sum_{i=1}^N \ell \rbr{\psi_\theta\rbr{R(i, L)}, Y_i} 
    \\ & = \E_{\rbr{R(i, L),  Y_i }\sim P_\gT^L} \sbr{ \ell \rbr{\psi_\theta(R(i, L) ), Y_i}  },
\end{align}
where each pair $\rbr{R(i, L), Y_i }$ can be viewed as an independent data point, and $P_\gT^L$ is the empirical distribution induced by these data points:
\begin{align}
    \cbr{\rbr{R(1, L), Y_1 }, \rbr{R(2, L), Y_2 },\cdots, \rbr{R(N, L), Y_N } }.
\end{align}
In such a way, we convert a graph $\gT$ to a distribution $P_\gT^L$. Based on this conversion, we can then make use of the distribution matching loss to synthesize another $\gS$ whose induced distribution $P_\gS^L$ is similar to $P_\gT^L$.

\subsubsection{Some Remarks.} 
The hop number $L$ has an effect on the size of the receptive field since it determines how many neighbours are included. Ideally, we want $L$ to be larger than the diameter of $\gT$, enabling the local graph to contain information of the entire graph. However, doing so will result in the following: 1) a large local graph and significant computational cost; 2) the introduction of noises 3) make the local graph indistinguishable, which is analogous to the over-smoothing problem in deep graph convolutional neural network \cite{DBLP:conf/iclr/KipfW17,DBLP:journals/corr/abs-1901-00596,DBLP:conf/aaai/LiHW18}. In many real-world datasets, using a larger number of layers does not increase the prediction accuracy even when training on the original large graph. 
In our studies, we will use a fixed $L$ during the condensation process, and then train the condensed graph with a $L'$ layer model where $L' \leq L$.

In this paper, we present GCDM under the context of node classification tasks. However, it is trivial to generalize the method to link prediction tasks.



\subsection{GCDM: Graph Condensation via Receptive Field Distribution Matching}
\label{sec:graph-condensation-with-distribution-matching}

With a graph dataset $\gT$ viewed as a distribution of receptive fields $P_\gT^L$, we aim to synthesize a small graph $\gS$ whose distribution of receptive fields $P_\gS^L$ is similar to  $P_\gT^L$.

\subsubsection{Synthesize the Labels \texorpdfstring{$Y'$}{Y'}}
Recall that the small graph $\mathcal{S} = \{A', X', Y'\}$ contains $A' \in \mathbb{R}^{N' \times N'}, X' \in \mathbb{R}^{N' \times d},$ and $ Y' \in \{0,...,C-1\}^{N'}$ with $N' \ll N$. To keep the class distribution of $Y'$ similar to that of $Y$, our method first determines the synthetic labels $ Y'$ by sampling
according to the class distribution in $Y$.

More precisely, let 
\begin{align}
    V_c:=\{i: Y_i = c\}
\end{align}
be the set of nodes that are in class $c$ in the original graph $\gT$. We sample the value of each synthetic label $Y'_j$ independently from the following categorical distribution:
\begin{align}
    P(Y_j' = c) = r_c:= \frac{|V_c|}{\sum_{c'=0}^{C-1} |V_{c'}|},\quad \text{for }c=1,\cdots, C-1,
    \label{eq:class-dist}
\end{align}
where $r_c$ denotes the class ratio. After the synthetic labels $Y'$ are sampled, we can define
\begin{align}
    V_c':=\{i: Y_i' = c\}
\end{align}
as the set of nodes that are in class $c$ in the synethetic graph $\gS$.

\subsubsection{Synthesize \texorpdfstring{$A'$}{A'} and \texorpdfstring{$X'$}{X'}.} Given the sampled labels $Y'$, we further optimize $A'$ and $X'$ to minimize the distance between distributions $P_\gS^L$ and  $P_\gT^L$.

To achieve the goal, within each class $c$, we optimize the maximum mean discrepancy (MMD)~ \cite{gretton2012kernel}  between the distributions of receptive fields according to the two graphs:

\begin{gather}
   \text{MMD}_c(\gT,\gS):= \nonumber 
   \sup_{\phi \in \gH} \left|\frac{1}{|V_c|}\sum_{i\in V_c} \phi\rbr{R_\gT(i,L)} -\frac{1}{|V_c'|}\sum_{j\in V_c'} \phi\rbr{R_\gS(j,L)} \right|,  
   \label{eq:mmd}
\end{gather}
where $\mathcal{H}$ is a family of functions, and we use the notation $R_\gT$ and $R_\gS$ to distinguish the receptive fields in different graphs $\gT$ and $\gS$. Note that when $\gH$ is a reproducing kernel Hilbert space, theoretical guarantees are available in the literature. By aggregating the MMD losses across all classes, the overall condensation loss is the following weighted sum:
\begin{align}
    \min_{A',X'} \sum_{c=1}^{C-1} r_c \cdot \text{MMD}_c(\gT,\gS). 
    \label{eq:overall-mmd}
\end{align}

\subsection{Parameterization and Algorithm}

In this section, we introduce how to optimize the condensation loss introduced in \eqref{eq:overall-mmd}. There are two main challenges. Firstly, we need to define the family of parametric function $\gH$ so that the MMD loss can be computed efficiently. Secondly, we need to parameterize the adjacency matrix $A'$ in a continuous way to avoid optimizing over a combinational solution space.

\subsubsection{Defining \texorpdfstring{$\gH$}{H} by GNNs} Due to the fact that (GNNs) inherently aggregate the information from each node's receptive field, they are ideal parametric functions $\phi\rbr{R(i,L)}$ for computing the statistics of receptive fields. Therefore, in GCDM, we define the function space $\gH$ via GNN models.

\subsubsection{Parametrize \texorpdfstring{$A'$}{A'} by MLP} As discussed in \cite{jin2021graph}, treating $A'$ and $X'$ as independent parameters ignores the inherent relationships between graph structure and node features, which have been well accepted in the literature. Therefore, we parameterize $A'$ as a function of the synthetic node features $X'$:
\begin{align}
    A' = g_\vartheta(X') \text{ with } A'_{ij} = \text{Sigmoid}\rbr{\text{MLP}_\vartheta([X_i'; X_j'])},
\end{align}
where $\text{MLP}_\vartheta$ is a multilayer perceptron, and $[\cdot;\cdot]$ denotes concatenation. With this parameterizaton, we optimize the parameters $\vartheta$ to find the synthetic $A'$.

\subsubsection{Condensation Loss in GCDM} With the function space $\gH$ defined by GNNs and the matrix $A'$ parameterized as a function of node features, we can now present the actual condensation loss used in GCDM. Let $\Phi_\theta$  be an $L$-layer GNN model parametrized by $\theta$, we solve the following optimization problem to generate the synthetic $X'$ and $A'$:
\begin{align}
    \min_{\vartheta, X'} \sum_{c=1}^{C-1} r_c \cdot \max_{\theta_c} \left\|\frac{1}{|V_c|}\sum_{i\in V_c} \emb_i^c - \frac{1}{|V_c'|}\sum_{j\in V_c'} \emb_j^c{}'\right\|_2^2 \label{eq:min-max}
\end{align}
where
\begin{align}
    \{\emb_i^c\}_{i=1}^N \leftarrow \Phi_{\theta_c}(A, X)\\
    \{\emb_j^c{}'\}_{j=1}^{N'} \leftarrow \Phi_{\theta_c}(A'=g_\vartheta(X'), X')
\end{align}
are node embeddings given by the GNN model $\Phi_\theta$. 

\subsubsection{Algorithm} To solve the optimization problem defined in \eqref{eq:min-max}, we adopt and algorithm that alternatively update the parameters $\vartheta, X'$ and $\theta_c$. The algorithm steps are lined out in Algorithm~\ref{algo:gcdm}. When implementing this algorithm, the gradient descent step can be replaced by other optimizers. 

\begin{algorithm}
    \KwIn{Training data $\mathcal{T}=(A, X, Y)$} 
    Obtain $Y'$ by sampling from the distribution in \eqref{eq:class-dist}\;
    Initialize $X'$ by randomly sample $N'$ node features from $X$\;
    Initialize $\vartheta, \{\theta_c\}_{c=0}^{C-1}$ randomly\;
    
    \For{$i = 1,\cdots, M$ } {
        \For{$e=1,\cdots, K_1$}{
            \For{$c=0,...,C-1$}{
                $A'\leftarrow g_\vartheta(X')$\;
                $\{\emb_i^c\}_{i=1}^N \leftarrow \Phi_{\theta_c}(A, X)$\;
                $\{\emb_j^c{}'\}_{j=1}^{N'} \leftarrow \Phi_{\theta_c}(A', X')$\;
                $\mathcal{L}\leftarrow r_c  \cdot \left\|\frac{1}{|V_c|}\sum_{i\in V_c} \emb_i^c - \frac{1}{|V_c'|}\sum_{j\in V_c'} \emb_j^c{}'\right\|_2^2$\;
                \uIf{$e\%(\tau_1+\tau_2) < \tau_1$}{
                Update $X' \leftarrow X' - \eta_1 \grad_{X'}\mathcal{L}$\;}
                \Else{
                Update $\vartheta   \leftarrow \vartheta - \eta_2 \grad_{\vartheta}\mathcal{L}$\;
                }
            }
        }
        \For{$e=1,\cdots, K_2$}{
            $A'\leftarrow g_\vartheta(X')$\;
            \For{$c=0,...,C-1$}{
                $\{\emb_i^c\}_{i=1}^N \leftarrow \Phi_{\theta_c}(A, X)$\;
                $\{\emb_j^c{}'\}_{j=1}^{N'} \leftarrow \Phi_{\theta_c}(A', X')$\;
                $\mathcal{L}\leftarrow  r_c \cdot \left\|\frac{1}{|V_c|}\sum_{i\in V_c} \emb_i^c - \frac{1}{|V_c'|}\sum_{j\in V_c'} \emb_j^c{}'\right\|_2^2$\;
                Update $\theta_c \leftarrow \theta_c + \eta_3 \grad_{\theta_c}\mathcal{L}$\;
            }
        }
     }
     $A' \leftarrow g_\vartheta(X')$\;
     $A'_{ij} \leftarrow 0$ if $A'_{ij} < 0.5$\;
     \KwOut{$\mathcal{S} = (A', X', Y')$\;}
     \caption{GCDM for Graph Condensation }
     \label{algo:gcdm}
\end{algorithm}

\subsubsection{A ``Graphless'' Variant: GCDM-X} Inspired by the work in \cite{jin2021graph}, we provide a model variant named GCDM-X that only producing synthetic node features $X'$ and fix the structure $A'$ to be an identity matrix $I$. Despite the fact that it does not learn the synthetic structure, this variation appears to be competitive in experiments. The potential reason could be that the node features are very informative, and the condensed $X'$ have incorporated relevant information from the graph.

\begin{table*}[h]
  \caption{Information used during condensation, training, and testing. X$'$ and A$'$ refer to the condensed graph.}
  \label{tab:info}
  \centering
  \begin{tabular}{lcccc}
    \toprule
    & GCOND-X & GCOND & GCDM-X & GCDM \\
    \midrule
    Condensation & A$_{train}$, X$_{train}$ & A$_{train}$, X$_{train}$ & A$_{train}$, X$_{train}$ & A$_{train}$, X$_{train}$\\
    Training & X$'$ & A$'$, X$'$ & X$'$ & A$'$, X$'$\\
    Test & A$_{test}$, X$_{test}$ & A$_{test}$, X$_{test}$ & A$_{test}$, X$_{test}$ & A$_{test}$, X$_{test}$\\
    \bottomrule
  \end{tabular}
\end{table*}

\section{EXPERIMENTS}

We conducted experiments on a range of graph datasets to evaluate the performance of GCDM. We were able to demonstrate the benefits of GCDM by comparing its performances to that of alternative baseline approaches, testing the generalizability of the condensed graphs on a variety of model architectures, running a speed comparison with the benchmark, and visualizing the condensed graphs. The information used in condensation, training and testing phase are demonstrated in \tabref{tab:info}.

\subsection{EXPERIMENTAL SETTINGS}

\textbf{Datasets.} We evaluate the performance of GCDM on five graph datasets: Cora \cite{sun2018graph}, PubMed \cite{xu2020building}, Citeseer \cite{DBLP:journals/corr/KipfW16}, Ogbn-arxiv \cite{hu2021open}, and Flikcr \cite{zeng2020graphsaint}. For all these datasets, we use public splits to split them into train, validation, and test sets. Dataset statistics are summarized in \tabref{tab:statistics}.
\newline
\textbf{Baselines.} We compare our proposed approach to six baseline methods: (i) one graph coarsening method \cite{huang2021scaling}, (ii-iv) three coreset methods (Random, Herding \cite{welling_2009}) and K-Center \cite{sener2018active}, (v) dataset condensation (DC) \cite{zhao2021datasetgm}, and (vi) the recent advance: Graph Condensation for Graph Neural Networks (GCOND) \cite{jin2021graph}. It should be noted that PubMed's benchmark coarsening results and the corresponding codes are not provided. Thus, we implement coarsening according to \cite{DBLP:journals/corr/abs-2106-05150}. 
\newline
\textbf{Experiment setting.} The experiment procedure consists of 3 steps: (1) given the training graph dataset, the condensation algorithm produces the condensed graph dataset; (2) train a GNN model using the condensed graph dataset and select the model using the original validation set; (3) evaluate the trained model on the original test dataset. Only the step (1) will be different for different condensation methods. After obtaining the condensed graph, the followed steps (2) and (3) are the same for all methods.

\begin{table*}[!ht]
  \caption{Dataset statistics.}
  \label{tab:statistics}
  \centering
  \begin{tabular}{lccccc}
    \toprule
    \bf{Dataset} & \bf{\#Nodes} & \bf{\#Edges} & \bf{\#Classes} & \bf{\#Features} & \bf{Training/Validation/Test}\\
    \midrule
    Cora & 2,708 & 5,429 & 7 & 1,433 & 140/500/1,000\\
    Pubmed & 19,717 & 88,648 & 3 & 500 & 60/500/1,000\\
    Citeseer & 3,327 & 4,732 & 6 & 3,703 & 120/500/1,000\\
    Flickr & 89,250 & 899,756 & 7 & 500 & 44,625/22,312/22,313\\
    Ogbn-arxiv & 169,343 & 1,166,243 & 40 & 128 & 90,941/29,799/48,603\\
    \bottomrule
  \end{tabular}
\end{table*}

\begin{table*}[!ht]
  \caption{Comparison of GCDM Performance with Baselines (Non-Empirical MMD)}
  \label{tab:performance-baseline}
  \noindent%
  \setlength\tabcolsep{0.01cm}
    \begin{tabular}{lccccccccccc}
    \toprule
    \multicolumn{2}{c}{} &\multicolumn{6}{c}{Baselines} & \multicolumn{2}{c}{Proposed} \\
    \cmidrule(lr){3-8} \cmidrule(lr){9-10} 
    Dataset &Ratio (\it r) & Random& Herding & K-Center & Coarsening & GCOND-X & GCOND &GCDM-X & GCDM & Whole\\
    & & (A', X') & (A', X') & (A', X')& (A', X')& (A')& (A', X') & (A')& (A', X') & Dataset  \\
    \midrule
    \multirow{3}{*}{Cora} 
    & 1.3\% & 63.6$\pm$3.7 & 67.0$\pm$1.3 & 64.0$\pm$2.3 & 31.2$\pm$0.2 & 75.9$\pm$1.2 & 79.8$\pm$1.3 & \textbf{81.3$\pm$0.4} & 69.4$\pm$1.3 & \multirow{3}{*}{82.5 $\pm$ 1.2}  \\
    & 2.6\% & 72.8$\pm$1.1 & 73.4$\pm$1.0 & 73.2$\pm$1.2 & 65.2$\pm$0.6 & 75.7$\pm$0.9 & 80.1$\pm$0.6 & \textbf{81.4$\pm$0.1} & 77.2$\pm$0.4\\
    & 5.2\% & 76.8$\pm$0.1 & 67.0$\pm$1.3 & 76.7$\pm$0.1 & 70.6$\pm$0.1 & 76.0$\pm$0.9 & 79.3$\pm$0.3 & \textbf{82.5$\pm$0.3} & 79.4$\pm$0.1\\
    \cmidrule(lr){1-11} 
    \multirow{3}{*}{Pubmed} 
    & 0.08\%  & 69.4$\pm$0.2 & \textbf{76.7$\pm$0.7} & 64.5$\pm$2.7 & 18.1$\pm$ $0.1^*$ & 69.1$\pm$1.1 & 76.5$\pm$0.2 & 75.5$\pm$0.3 & 75.7$\pm$0.3 &\multirow{3}{*}{79.3 $\pm$ 0.2}\\
    & 0.15\%  & 73.3$\pm$0.7 & 76.2$\pm$0.5 & 69.4$\pm$0.7 & 28.7$\pm$ $4.1^*$ & 73.2$\pm$0.7 & 77.1$\pm$0.5 & 75.7$\pm$0.3 & \textbf{77.3$\pm$0.1} & \\
    & 0.3\%  & 77.8$\pm$0.3 & 78.0$\pm$0.5 & 78.2$\pm$0.4 & 42.8$\pm$ $4.1^*$ & 71.7$\pm$0.9 & 77.9$\pm$0.4 & 77.2$\pm$0.2 & \textbf{78.3$\pm$0.3} & \\
    \cmidrule(lr){1-11} 
    \multirow{3}{*}{Citeseer} 
    & 0.9\% & 54.4$\pm$4.4 & 57.1$\pm$1.5 & 52.4$\pm$2.8 & 52.2$\pm$0.4 & \textbf{71.4$\pm$0.8} & 70.5$\pm$1.2 & 69.0$\pm$0.5 & 62.0$\pm$0.1 & \multirow{3}{*}{73.0 $\pm$ 0.1} \\
    & 1.8\% & 64.2$\pm$1.7 & 66.7$\pm$1.0 & 64.3$\pm$1.0 & 59.0$\pm$0.5 & 69.8$\pm$1.1 & 70.6$\pm$0.9 & \textbf{71.9$\pm$0.5} & 69.5$\pm$1.1\\
    & 3.6\% & 69.1$\pm$0.1 & 69.0$\pm$0.1 & 69.1$\pm$0.1 & 65.3$\pm$0.5 & 69.4$\pm$1.4 & 69.8$\pm$1.4 & \textbf{72.8$\pm$0.6} & 69.8$\pm$0.2\\
    \cmidrule(lr){1-11}
    \multirow{3}{*}{Flickr}
    & 0.1\% & 41.8$\pm$2.0 & 42.5$\pm$1.8 & 42.0$\pm$0.7 & 41.9$\pm$0.2 & 45.9$\pm$0.1 & 46.5$\pm$0.4 & 46.0$\pm$0.1 & \textbf{46.8$\pm$0.2} & \multirow{3}{*}{50.2 $\pm$ 0.3}  \\
    & 0.5\% & 44.0$\pm$0.4 & 43.9$\pm$0.9 & 43.2$\pm$0.1 & 44.5$\pm$0.1 & 45.0$\pm$0.2 & 47.1$\pm$0.1 & 47.4$\pm$0.3 & \textbf{47.9$\pm$0.3} & \\
    & 1\%   & 44.6$\pm$0.2 & 44.4$\pm$0.6 & 44.1$\pm$0.4 & 44.6$\pm$0.1 & 45.0$\pm$0.1 & 47.1$\pm$0.1 & 47.1$\pm$0.2 & \textbf{47.5$\pm$0.1} &\\
    \cmidrule(lr){1-11} 
    \multirow{2}{*}{Ogbn-arxiv} 
    & 0.25\% & 57.3$\pm$1.1 & 58.6$\pm$1.2 & 56.8$\pm$0.8 & 43.5$\pm$0.2 & \textbf{64.2$\pm$0.4} & 63.2$\pm$0.3 & 61.2$\pm$0.1 & 59.6$\pm$0.4 & \multirow{2}{*}{71.4 $\pm$ 0.1}\\
    & 0.5\% & 60.0$\pm$0.9 & 60.4$\pm$0.8 & 60.3$\pm$0.4 & 50.4$\pm$0.1 & 63.1$\pm$0.5 & \textbf{64.0$\pm$0.4} & 62.5$\pm$0.1 & 62.4$\pm$0.1 & \\
     
    \bottomrule
  \end{tabular}
\end{table*}

\subsection{Test Accuracy Comparison}

To evaluate whether the GNN trained on the condensed graph can perform well on the test set in the original graph, we report the test accuracy for each method in \tabref{tab:performance-baseline}. The results in this table are achieved by using a two-layer GCN to perform training and testing after the condensation procedure. For the GCOND benchmark \cite{jin2021graph}, the GNNs utlized during condensation vary according to the datasets, with the authors making the decisions; For GCDM, we use a two-layer GCN while condensing under the GCDM-X framework and a two-layer SGC while condensing under the GCDM framework. The latter SGC was chosen to conserve GPU memory and improve the speed of condensing.

\begin{table*}[!ht]
  \caption{Graph generated by Receptive Field Distribution Matching using a single condense architecture could generalize to other architectures in evaluation, while the performance is comparable or superior to GCOND.}
  \label{tab:gen}
  \centering
  \setlength\tabcolsep{4pt}
  \begin{tabular}{lcccccccccc}
    \toprule
     &Methods & Data& GIN &MLP & APPNP & Cheby & GCN & SAGE & SGC & Avg.\\
    \midrule
    \multirow{4}{*}{\shortstack{Flickr\\ \it r \it= \it 0.5\%}} & GCOND-X & X'& 29.5 & 41.5 & 44.1 & 31.9 &  47.6 & 32.6 & 47.7 & 39.3\\
    & GCOND & A',X'&  38.8 & 42.2 & 45.3 & 34.9 & 45.8 & 43.5 & 46.2 & 42.4\\
    & GCDM-X & X' & 42.2 & 42.3 & 42.4 & 42.4 & 47.4 & 42.6 & 45.8 & 43.6 \\
    & GCDM & A',X' & 42.5 & 43.9 & 45.9 & 43.2 & 47.8 & 42.9 & 49.4 & 45.1 \\
    \cmidrule(lr){1-11} 
    \multirow{4}{*}{\shortstack{Ogbn-arxiv\\ \it r \it= \it 0.5\%}} & GCOND-X & X' & 59.9 & 47.6 & 55.2 & 47.4 & 62.9 & 59.3 & 64.7 & 56.7\\
    & GCOND & A',X' & 7.2* & 46.4 & 55.1 & 44.7 & 64.7 & 37.8 & 64.8 & 47.8\\
    & GCDM-X & X' &58.6 & 44.5 & 52.8 & 45.1 & 62.5 & 57.4 & 64.1 & 55.0 \\
    & GCDM & A',X' &58.1 & 45.0 & 52.6 & 45.1 & 62.0 & 56.8 & 62.2 & 54.5 \\
    
    \bottomrule
  \end{tabular}
\end{table*}


\tabref{tab:performance-baseline} summarizes the node classification performance, from which the following observations are made:

{\bfseries Obs 1. GCDM and GCDM-X achieve promising performance with high reduction rates.} Deep learning based condensation methods like GCOND and GCDM outperform coreset selection methods by a large margin. Our methods achieve  81.3\%, 81.4\%, and 82.5\% accuracies at 1.3\%, 2.6\% and 5.2\% condensation ratios on Cora dataset, while the model trained on the whole graph achieves an accuracy of 82.5\%. Our methods also show promising performance on the other four datasets. Additionally, for most settings, our framework outperforms the current SOTA framework: GCOND.

{\bfseries Obs 2. Learning X$'$ solely can also result in good performance.}
Similar to what was reported in GCOND, we observed that GCDM-X often achieves close performance to GCDM. In some cases, GCDM-X even works better. The reason could be that $X'$ has already encoded node features and structural information of the original graph during the condensation process, or that the node features are very informative in real-world datasets.


{\bfseries Obs 3. Larger sample size leads to improved performance.}
The authors of GCOND \cite{jin2021graph} reported that increasing the size of the condensed graph doesn't necessarily improve performance. It is stated that once the condensation ratio reaches a certain value, the performance stops improving. In our experiments, we find the performance improves as the size of the condensed graph increases. This may attribute to the inherent property of distribution matching. Larger graph sizes allow for more sample points to mimic the distribution of the original graph. When the distribution of the original graph becomes more complicated, additional sample points may be required to approximate the distribution. It enables the synthetic graph generated by our method to encapsulate additional information from the original graph. Notably, our method under-performs GCOND on Ogbn-arxiv with the same condensation ratio. However, when we run a experiment with a 1\% condensation rate, we obtain a test accuracy of \textbf{64.9\%}.

\subsection{GENERALIZED PERFORMANCE}
In this section, we determine whether or not a graph condensed using a particular GNN architecture can generalize to other GNN models during the evaluation phase. 
For the experiments in this section, we use GCN during the condensation procedure for both GCOND and GCDM. After the condensed graphs are obtained, we perform training and testing using various models including: GIN \cite{gin2018}, MLP \cite{hu2021graphmlp}, APPNP \cite{klicpera2019predict}, Cheby \cite{tang2019chebnet}, GCN \cite{DBLP:journals/corr/KipfW16}, SAGE \cite{hamilton2018inductive}, and SGC \cite{wu2019simplifying}. The test accuracies on the two largest datasets, Flickr and ogbn-Arxiv, are summarized in \tabref{tab:gen}. We can observe consistent improvement in comparison to GCOND, most notably when SAGE and GIN are used. One reason that GCDM outperforms GCOND in terms of generalization ability is that GCOND utilizes gradients during a single model training process, which may not translate well to another training process. It should be noted that the abnormal GIN performance in the table was obtained by execution of the original GCOND program.

\begin{table*}[!ht]
  \caption{Some condensed graph could produce more visually clustered results compared to GCOND. NPC: Node Per Class.}
\begin{center}
\setlength{\tabcolsep}{0.001cm}
\begin{tabular}{ccccc}
\includegraphics[width=0.18\linewidth]{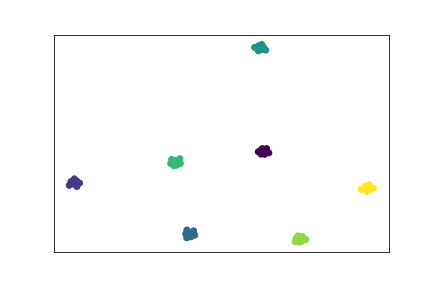} &
\includegraphics[width=0.18\linewidth]{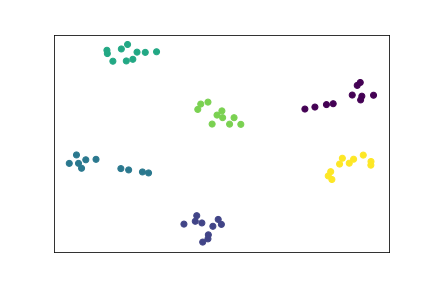} &
\includegraphics[width=0.18\linewidth]{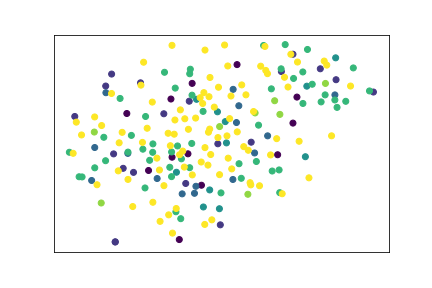} &
\includegraphics[width=0.18\linewidth]{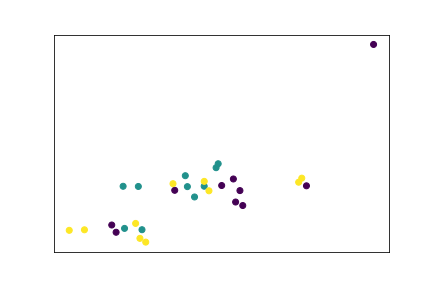} &
\includegraphics[width=0.18\linewidth]{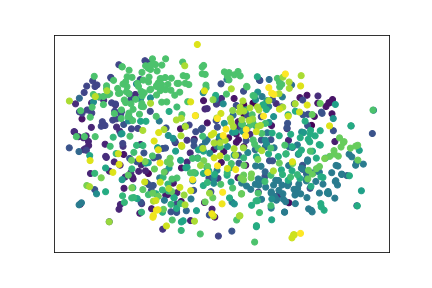} \\
 (a) Cora 10 NPC &  (b) Citeseer 10 NPC  & (c) Flickr 30 NPC & (d) Pubmed 10 NPC &  (e) ogbn-Arxiv 20 NPC\\
\end{tabular}
\caption{Some condensed graph could produce more visually clustered results compared to GCOND. NPC: Node Per Class.}
\label{tab:visualization}
\end{center}
\end{table*}

\begin{table*}[!ht]
  \caption{Running time comparison between GCDM and GCOND on a single RTX8000 GPU (50 epochs).}
  \label{tab:speedperformance}
  \centering
  \begin{tabular}{lcccccccc}
    \toprule
    Dataset & Methods &  0.1\% & 0.5\% & 1\%\\
    \midrule
    \multirow{2}{*}{Ogbn-arxiv}& 
    GCOND (50 epochs) & 1083.3s & 2028.5s & 5960.4s\\
    & GCDM (50 epochs) & 981.8s & 1125.6s & 1781.8s\\
    \bottomrule
  \end{tabular}
\end{table*}

\subsection{SPEED PERFORMANCE}
This section examines and compares the condensing speeds of GCOND and GCDM. The overall condensation algorithm of GCOND (see Algorithm 1 in \cite{jin2021graph}) has a very similar structure to GCDM (Algorithm~\ref{algo:gcdm}) except that their condensation losses are different. Therefore, we report the time required to run 50 epochs ($M=50$ in Algorithm~\ref{algo:gcdm}) on the ogbn-arxiv dataset for both GCDM and GCOND in \tabref{tab:speedperformance}. Note that to achieve the performances reported in \tabref{tab:performance-baseline}, GCOND needs to run for around 200 epochs, and GCDM runs for less than 150 epochs. As the size of the condensed graph increases, the time required for GCOND increases considerably faster than that of GCDM.

\subsection{VISUALIZATION OF CONDENSED GRAPHS}
We utilize T-SNE \cite{JMLR:v9:vandermaaten08a} to embed the the condensed graph's node features in two dimensions and present the resulting visualizations in \tabref{tab:visualization}. As shown, the condensed node features occasionally exhibit a more clustered pattern compared to the visualizations of GCOND \cite{jin2021graph}, as demonstrated by the condensed nodes for Cora, CiteSeer, and Pubmed. While the patterns of other datasets such as Flickr and Ogbn-arxiv are not as immediately discernible, similar to the visualizations presented by GCOND.


\section{CONCLUSION}
In this paper, we investigate a novel methodology called GCDM for generating a small synthetic graph from a large and complex graph. The GCDM framework is implemented by optimizing the synthetic graph using a distribution matching loss measured by maximum mean discrepancy (MMD). Through the proposed framework, we are able to obtain comparable evaluation performance to the original graph. The synthetic graph is also generalizable to a variety of downstream models. This enables neural architecture search and hyperparameter tuning to be done in a highly efficient way. Additionally, the improvement in condensation speed is demonstrated, allowing greater flexibility in a model-retraining setting.

\appendix
\bibliographystyle{unsrt}  
\bibliography{bibfile}
\end{document}